\documentclass[10pt, a4paper]{article}
\usepackage{lrec2000}
\usepackage{alltt,epsfig,url}
\usepackage{endnotes}\renewcommand{\footnote}{\endnote}
\def\myurl#1{{\small[\url{#1}]}}
\def\smtt#1{{\small\tt #1}}

\renewenvironment{quote}
                 {\list{}{\small\leftmargin\parindent}%
                 \item\relax}
                 {\endlist}

\title{Seven Dimensions of Portability for\\
Language Documentation and Description}

\name{Steven Bird$^{\ast}$ and Gary Simons$^{\dagger}$} 

\address{
$^{\ast}$Linguistic Data Consortium, University of Pennsylvania,
3615 Market Street, Philadelphia, PA 19104, USA\\
$^{\dagger}$SIL International, 7500 West Camp Wisdom Road, Dallas, TX 75236, USA
}

\abstract{The process of documenting and describing the world's languages
is undergoing radical transformation with the rapid uptake of new
digital technologies for capture, storage, annotation and
dissemination.  However, uncritical adoption of new tools and
technologies is leading to resources that are difficult to reuse and
which are less portable than the conventional printed resources they
replace.  We begin by reviewing current uses of software tools and
digital technologies for language documentation and description.  This
sheds light on how digital language documentation and description are
created and managed, leading to an analysis of seven portability
problems under the following headings: content, format, discovery,
access, citation, preservation and rights.  After characterizing each
problem we provide a series of value statements, and this provides the
framework for a broad range of best practice
recommendations.}

\begin{document}

\maketitleabstract

\section{Introduction}

It is now easy to collect vast quantities of language documentation and
description and store it in digital form.  It is getting easier to
transcribe the material and link it to linguistic descriptions.  Yet how
can we ensure that such material can be re-used by others, both now and
into the future?  While today's linguists can access documentation that is over
100 years old, much digital language documentation and description
is unusable within a decade of its creation.

The fragility of digital records is amply demonstrated.  For
example, the interactive video disks created by the BBC Domesday Project
are inaccessible just 15 years after their creation.\footnote{%
\url{http://www.observer.co.uk/uk_news/story/0,6903,661093,00.html}} In the
same way, linguists who are quick to embrace new technologies and create
digital materials in the absence of archival formats and practices soon
find themselves in technological quicksand.

The uncritical uptake of new tools and technologies is encouraged by
sponsors who favor projects that promise to publish their data on the
web with a search interface.  However, these projects depend on
technologies with life cycle of 3-5 years, and the resources
they create usually do not outlive the project any longer than this.

This paper considers portability in the broadest sense:
across different software and hardware platforms;
across different scholarly communities (e.g.\ field linguistics, language technology);
across different purposes (e.g.\ research, teaching, development);
and across time.  Portability is frequently treated as an issue for software,
but here we will focus on data.  In particular,
we address portability for language documentation
and description, and interpret these terms following Himmelmann:

{\small
\begin{quote}
The aim of a language documentation is to provide a comprehensive
record of the linguistic practices characteristic of a given speech
community. Linguistic practices and traditions are manifest in two
ways: (1) the observable linguistic behavior, manifest in everyday
interaction between members of the speech community, and (2) the
native speakers' metalinguistic knowledge, manifest in their ability
to provide interpretations and systematizations for linguistic units
and events. This definition of the aim of a language documentation
differs fundamentally from the aim of language descriptions: a
language description aims at the record of A LANGUAGE, with "language"
being understood as a system of abstract elements, constructions, and
rules that constitute the invariant underlying structure of the
utterances observable in a speech community. \cite[166]{Himmelmann98}
\end{quote}}

We adopt the cover term DATA to mean any information that documents
or describes a language, such as a published monograph, a computer
data file, or even a shoebox full of hand-written index cards. The
information could range in content from unanalyzed sound recordings to
fully transcribed and annotated texts to a complete descriptive
grammar.  Beyond data, we are be concerned with language resources
more generally, including tools and advice.
By TOOLS we mean computational resources that facilitate
creating, viewing, querying, or otherwise using language data. Tools
include software programs, along with the digital resources
that they depend on such as fonts, stylesheets, and document
type definitions.  By ADVICE we mean any information about what data
sources are reliable, what tools are appropriate in a given situation,
and what practices to follow when creating new data
\cite{BirdSimons01}.

This paper addresses seven dimensions of portability for digital
language documentation and description, identifying problems,
establishing core values, and proposing best practices.
The paper begins with a survey of the tools and
technologies (\S2), leading to a discussion of the problems that arise
with the resources created using these tools and
technologies (\S3).  We identify seven kinds of portability problem,
under the headings of content, format, discovery, access, citation,
preservation and rights.  Next we give statements about core values in
digital language documentation and description, leading to a series of
``value statements'', or requirements for best practices (\S4), and
followed up with collection of best practice recommendations (\S5).
The structure of the paper is designed to build consensus.  For
instance, readers who take issue with a best practice recommendation
in \S5 are encouraged to review the corresponding statement of values
in \S4 and either suggest a different practice which better implements
the values, or else take issue with the value statement (then back up
to the corresponding problem statement in \S3, and so forth).

\section{Tools and Technologies for Language Documentation and Description}

Language documentation projects are increasing in their reliance on
new digital technologies and software tools.  This section contains a
comprehensive survey of the range of practice, covering general
purpose software, specialized tools, and digital technologies.
Reviewing the available tools gives us a snapshot of how digital
language documentation and description is created and managed, and
provides a backdrop for our analysis of data portability problems.

\subsection{General purpose tools}

The most widespread practice in language documentation involves the
use of office software.  This software is readily available, often
pre-installed, and familiar.  Word processors
have often been used as the primary storage for large lexical
database, including a Yoruba lexicon with 30,000 entries split
across 20 files.
Frequently cited benefits are the \textsc{wysiwyg} editing, the find/replace
function, the possibility of cut-and-paste to create sublexicons,
and the ease of publishing.
Of course, a large fraction of the linguist's time is spent on
maintaining consistency across multiple copies of the same data.
Word processors have also been used
for interlinear text, with three main approaches: fixed width fonts
with hard spacing, manual setting of tabstops, and tables.\footnote{%
\url{http://www.linguistics.ucsb.edu/faculty/cumming/WordForLinguists/Interlinear.htm}}
All methods require manual line-breaking, and significant labor if
line width or point size are changed.  Another kind of office software is
the spreadsheet, which is often used for wordlists.
Language documentation created using office software is normally
stored in a secret proprietary format that is unsupported within 5-10
years.  While other export formats are supported, they may loose
some of the structure.  For instance, part of speech may be
distinguished in a lexical entry through the use of a particular font,
and this information may be lost when the data is exported.
Also, the portability of export formats may be
compromised, by being laden with presentational markup.

A second class of general purpose software is the hypertext processors.
Perhaps the first well-known application to language documentation
was the original Macintosh hypercard stacks of {\it Sounds of the World's Languages}
\cite{Ladefoged96}.  While it was easy to create a complex web of navigable pages,
nothing could overcome the limitations of a vendor-specific hypertext language.
More recently, the \textsc{html} standard and universal, free browsers have encouraged
the creation of large amounts of hypertext for a variety of
documentation types.  For instance, we have
interlinear text with \textsc{html} tables
(e.g.\ Austin's Jiwarli fieldwork\footnote{%
\url{http://www.linguistics.unimelb.edu.au/research/projects/jiwarli/gloss.html}}),
interlinear text with \textsc{html} frames
(e.g.\ Culley's presentation of Apache texts\footnote{%
\url{http://etext.lib.virginia.edu/apache/ChiMesc2.html}}),
\textsc{html} markup for lexicons, with hyperlinks from glossed examples and a thesaurus
(e.g.\ Austin and Nathan's Gamilaraay lexicon\footnote{%
\url{http://www3.aa.tufs.ac.jp/~austin/GAMIL.HTML}}),
gifs for representing IPA transcriptions
(e.g.\ Bird's description of tone in Dschang\footnote{%
\url{http://www.ldc.upenn.edu/sb/fieldwork/}}),
and Javascript for image annotations
(e.g.\ Poser's annotated photographs of gravestones
engraved with D\'en\'e syllabics\footnote{%
\url{http://www.cnc.bc.ca/yinkadene/dakinfo/dulktop.htm}}).
In all these cases, \textsc{html} is used as the primary storage format, not simply as
a view on an underlying database.  The intertwining of content and
format makes this kind of language documentation difficult to maintain and re-use.

The third category of general purpose software is database packages.
In the simplest case, the creator shares the database with others by requiring
them to purchase the same package, and by shipping them a full dump of the database
(e.g.\ the StressTyp database, which requires users to buy a copy of
``4th Dimension''\footnote{%
\url{http://fonetiek-6.leidenuniv.nl/pil/stresstyp/stresstyp.html}}).
A more popular approach is to put the database on a web-server, and create
a forms-based web interface that allows remote users to search the database
without installing any software
(e.g.\ the Comparative Bantu Online Lexical Database\footnote{%
\url{http://www.linguistics.berkeley.edu/CBOLD/}}
and the Maliseet-Passamaquoddy Dictionary.\footnote{%
\url{http://ultratext.hil.unb.ca/Texts/Maliseet/dictionary/index.html}})
Recently, some sites have started allowing database updates via the web
(e.g.\ the Berkeley Interlinear Text Collector\footnote{%
\url{http://ingush.berkeley.edu:7012/BITC.html}}
and the Rosetta Project's site for uploading texts, wordlists and descriptions\footnote{%
\url{http://www.rosettaproject.org:8080/live/}}).

\subsection{Specialized tools}

Over the last two decades, several dozen tools have been developed
having specialized support for language documentation and description.
We list a representative sample here; more can be found on SIL's
page on \textit{Linguistic Computing Resources},\footnote{%
\url{http://www.sil.org/linguistics/computing.html}} on the
\textit{Linguistic Exploration} page,\footnote{%
\url{http://www.ldc.upenn.edu/exploration/}}
and on the \textit{Linguistic Annotation} page.\footnote{%
\url{http://www.ldc.upenn.edu/annotation/}}

Tools for linguistic data management include
Shoebox\footnote{\url{http://www.sil.org/computing/shoebox/}}
and the Fieldworks Data Notebook.\footnote{\url{http://fieldworks.sil.org/}}
Speech analysis tools include Praat\footnote{\url{http://fonsg3.hum.uva.nl/praat/}}
and SpeechAnalyzer.\footnote{\url{http://www.sil.org/computing/speechtools/speechanalyzier.htm}}
Many specialized signal annotation tools have been developed, including
CLAN,\footnote{\url{http://childes.psy.cmu.edu/}}
EMU,\footnote{\url{http://www.shlrc.mq.edu.au/emu/}}
TableTrans, InterTrans, TreeTrans.\footnote{\url{http://sf.net/projects/agtk/}}
There are many orthographic transcription tools, including
Transcriber\footnote{\url{http://www.etca.fr/CTA/gip/Projets/Transcriber/}}
and MultiTrans.\footnote{\url{http://sf.net/projects/agtk/}}
There are morphological analysis tools, such as the
Xerox finite state toolkit.\footnote{\url{http://www.xrce.xerox.com/research/mltt/fst/}}
There are a wealth of concordance tools.
Finally, some integrated multi-function systems have been created, such as
LinguaLinks Linguistics Workshop.\footnote{\url{http://www.sil.org/LinguaLinks/LingWksh.html}}

In order to do their specialized linguistic processing, each of these tools
depends on some model of linguistic information.  Time-aligned
transcriptions, interlinear texts, syntax trees, lexicons, and so forth,
all require suitable data structures and file formats.  Given that most of
these tools have been developed in isolation, they typically employ
incompatible models and formats.  For example, data created with an
interlinear text tool cannot be subsequently annotated with syntactic
information without losing the interlinear annotations.  When interfaces
and formats are open and documented, it is occasionally possible to cobble
the tools together in support of a more complex need.  However, the result
is a series of increasingly baroque and decreasingly portable approximations
to the desired solution.
Computational support for language documentation and description is in disarray.

\subsection{Digital technologies}

A variety of digital technologies are now used in language
documentation thanks to sharply declining hardware costs.  These
include technologies for digital signal capture (audio, video,
physiological) and signal storage (hard disk, \textsc{cd-r}, \textsc{dvd-r}, minidisc).
Software technologies are also playing an influential role as new
standards are agreed.  The most elementary and pervasive of these is
the hyperlink, which makes it possible to connect linguistic
descriptions to the underlying documentation (e.g.\ from an analytical
transcription to a recording).  Such links streamline the descriptive
process; checking a transcription can be done with mouse clicks
instead of digging out a tape or finding an informant.  The ability to
navigate from description to documentation also facilitates analysis
and verification.  Software technologies and standards have given rise
to the internet which permits low-cost dissemination of language
resources.  Notably, it is portability problems with these tools and
formats that prevents these basic digital technologies from having
their full impact.  The download
instructions for the Sumerian lexicon\footnote{%
\url{http://www.sumerian.org/}}
typify the problems (hyperlinks are underlined):

\begin{quote}
\scriptsize
\underline{
Download the Sumerian Lexicon as an Adobe Acrobat PDF file.} In order to
minimize downloads of this large file, once you have it, please use your
Acrobat Reader to save it and retrieve it to and from your own desktop.

\underline{
Download the Sumerian Lexicon as a Word for Windows 6.0 file in a self-}
\underline{extracting WinZip archive.}

\underline{Download the same contents in a non-executable zip file.}

Includes version 2 of the Sumerian True Type font for displaying
transliterated Sumerian. Add the font to your installed Windows fonts at
Start, Settings, Control Panel, Fonts. To add the Sumerian font to your
installed Windows fonts, you select File and Add New Font. Afterwards, make
sure that when you scroll down in the Fonts listbox, it lists the Sumerian
font. When you open the SUMERIAN.DOC file, ensure that at File, Templates,
or at Tools, Templates and Add-Ins, there is a valid path to the enclosed
SUMERIAN.DOT template file.
If you do not have Microsoft's Word for Windows, you can download a
free Word for Windows viewer at \underline{Microsoft's Web Site.}

\underline{
Download Macintosh utility UnZip2.0.1} to uncompress IBM ZIP files. To
download and save this file, you should have Netscape set in Options,
General Preferences, Helpers to handle hqx files as Save to Disk. Decode
this compressed file using Stuffit Expander.
\underline{
Download Macintosh utility TTconverter} to convert the IBM format
SUMERIAN.TTF TrueType font to a System 7 TrueType font. Decode this
compressed file using Stuffit. Microsoft Word for the Macintosh can read a
Word for Windows 6.0 document file. There is no free Word for Macintosh
viewer, however.
\end{quote}

\subsection{Digital Archives}

Recently several digital archives of language documentation and description have
sprung up, such as the
Archive of the Indigenous Languages of Latin America,\footnote{%
\url{http://www.ailla.org/}}
and the Rosetta Project's Archive of 1000 Languages.\footnote{%
\url{http://www.rosettaproject.org/}}
These exist alongside older archives which are in various stages
of digitizing their holdings:
the Archive of the Alaska Native Language Center,\footnote{%
\url{http://www.uaf.edu/anlc/}}
the LACITO Linguistic Data Archive,\footnote{%
\url{http://195.83.92.32/index.html.en}}
and the US National Anthropological Archives.\footnote{%
\url{http://www.nmnh.si.edu/naa/}}
These archives and many others are surveyed on the
\textit{Language Archives} page.\footnote{%
\url{http://www.ldc.upenn.edu/exploration/archives.html}}
Under the aegis of \textsc{olac},
the \textit{Open Language Archives Community},\footnote{%
\url{http://www.language-archives.org/}}
the notion of language archive has been broadened to include archives
of linguistic software, such as the
Natural Language Software Registry\footnote{%
\url{http://registry.dfki.de/}}

These archives face many challenges, the most significant being the
lack of funding.  Other challenges may include:
identifying, adapting and deploying digital archiving standards;
setting up key operational functions such as offsite backup,
migration to new digital formats and media over time, and the support of new
access modes (e.g.\ search facilities) and delivery formats
(e.g.\ streaming media); and
obtaining the long-term support of a major institution to assure
contributors and users that the materials will be available over
the long term.

\section{Seven Problems for Portability}

With the rapid uptake of new digital technologies, many creators of
language documentation and description are ignoring the question of
portability, with the unfortunate consequence that the fruits of their
labors are likely to be unusable within 5-10 years.  In this section we discuss
seven critical problems for the portability of this data.

\subsection{Content}

Many potential users of language data are
interested in assimilating multiple descriptions of a single language
to gain an understanding of the language which is as comprehensive as
possible.  Many users are interested in comparing the descriptions of
different languages in order to apply insights from one analysis to
another or to test a typological generalization.  However, two
descriptions may be difficult to compare or assimilate because they
have used terminology differently, or because the documentation on which
the descriptions are based is unavailable.

Language documentation and description of
all types depends critically on technical vocabulary, and
ambiguous terms compromise portability.  For instance, the symbols
used in phonetic transcription have variable interpretation depending
on the descriptive tradition: ``it is crucial to be aware of the
background of the writer when interpreting an unexplained occurrence
of [y]'' \cite[168]{PullumLadusaw86}.
In morphosyntax, the term ``absolutive'' can refer to one of the cases
in an ergative language, or to the unpossessed form of a noun (in the
Uto-Aztecan tradition) \cite[151]{Lewis01},
and a correct interpretation of the term depends on an understanding of
the linguistic context.

This terminological variability leads to problems for retrieval.  Suppose
that a linguist wanted to search the full-text content of a large
collection of digital language data, in order to discover which other languages have
relevant phenomena.  Since there are no standard ontologies, the user
will discover irrelevant documents (low precision) and will fail
to discover relevant documents (low recall).  In order to carry out a
comprehensive search, the user must know all the ways in which
a particular phenomena is described.  Even once a set of descriptions are retrieved,
it will generally not be possible to draw reliable
comparisons between the descriptions of different languages.

The content of two descriptions may also be difficult to reconcile
because it is not possible to verify them with respect to the language documentation
that they cite.  For example, when two descriptions of the same language provide
different phonetic transcriptions of the same word, is this the result of a
typographical error, a difference in transcription practice, or a genuine difference
between two speech varieties?  When two descriptions of different
languages report that the segmental inventories of both languages contain a [k],
what safe conclusions can be drawn about how similar the two sounds are?
Since the underlying documentation is not available,
such questions cannot be resolved, making it difficult to re-use the resources.

While the large-scale creation of digital language resources is a recent
phenomenon, the language documentation community has been active since
the 19th century, and much earlier in some instances.  At risk of oversimplifying,
a widespread practice over this extended period has been to collect wordlists
and texts and to write descriptive grammars.  With the arrival of new digital
technologies it is easy to transfer the whole endeavor from paper to computer,
and from tape recorder to hard disk, and to carry on just as before.  Thus,
new technologies simply provide a better way to generate the old kinds of resources.
Of course this is a wasted opportunity, since the new technologies can also
be used to create digital multimedia recordings of rich linguistic events.
Such rich recordings often capture items which turn out to be useful in
later linguistic analysis, and have
immense intrinsic value as a record of cultural heritage for
future generations.  However, managing digital technologies in less controlled situations
leads to many technical and logistical issues, and there are no guidelines for integrating new
technologies into new documentary practices.

\subsection{Format}

Language data frequently ends up in a secret proprietary format using
a non-standard character encoding.  To use such data one must often
purchase commercial software then install it on the same hardware and
under the same operating system used by the creator of the data.

Other formats, while readable outside the tool that created them,
remain non-portable when they are not explicitly documented.  For
example, the interpretation of the field names in Shoebox format may
not be documented, or the documentation may become separated from the
data file, making it difficult to guess what the different fields
signify.

The developers of linguistic tools must frequently parse presentational
formats.  For example, the occurrence of \smtt{<b>[n]</b>} in a lexical
entry might indicate that this is an entry for a noun.  More difficult cases
involve subtle context-dependencies.  This presentational markup
obscures the structure and interpretation of the linguistic content.
Conversely, in the absence of suitable browsing and rendering tools,
end-users must attempt to parse formats that were designed
to be read only by machines.

\subsection{Discovery}

Digital language data is often presented as a physical or digital
artefact with no external description.  Like a book without a cover
page or a binary file called {\small\tt dict.dat}, one is forced to
expend considerable effort to discover the subject matter and the
nature of the content.  Organized collections -- such as the archive
of a university linguistics department -- may provide some
metadescription, but it is likely to use a parochial format and
idiosyncratic descriptors.  If they are provided, key descriptors like
{\em subject language} and {\em linguistic type} are usually given in
free text rather than a controlled vocabulary, reducing precision and
recall.  As a consequence, discovering relevant language resources is
extremely difficult, and depends primarily on word-of-mouth and
queries posted to electronic mailing lists.  Thus, new
resource creation efforts may proceed in ignorance of prior and
concurrent efforts, wasting scarce human resources.

In some cases, one may obtain a resource only to discover upon closer
inspection that it is in an incompatible format.  This is the
flip-side of the discovery problem.  Not only do we need to know that
a resource exists, but also that it is relevant.  When resources are
inadequately described, it is difficult (and often impossible) to find
a relevant resource, a huge impediment to portability.

\subsection{Access}

In the past, primary documentation was usually not disseminated.  To listen
to a field recording it was often necessary to visit the laboratory of the person
who collected the materials, or to make special arrangements for the
materials to be copied and posted.  Digital publication on the web has
alleviated this problem, although projects usually refrain from full
dissemination by limiting access via a restrictive search interface.
This means that only selected portions of the documentation can be
downloaded, and that all access must use categories predefined by the
provider.  Moreover, these web forms only have a lifespan of 3-5 years,
relying on ad hoc CGI scripts which may cease working when the interpreter
or webserver are upgraded.  Lack of full access means that materials are not
portable.
More generally, people have often conflated digital publication with
web publication, and publish high-bandwidth materials on the web which
would be more usable if published on \textsc{cd} or \textsc{dvd}.

Many language resources have applications beyond those envisaged by their
creators.  For instance, the Switchboard database \cite{Godfrey92}, collected
for the development of speaker-independent automatic speech recognition, has
since been used for studies of intonation and disfluency.  Often this redeployment
is prevented through the choice of formats.  For instance, publishing
conversation transcripts in the Hub-4 \textsc{sgml} format does not facilitate their
reuse in, say, conversational analysis.  In other cases, redeployment is
prevented by the choice of media.  For instance, an endangered language
dictionary published only on the web will not be accessible to speakers of
that language who live in a village without electricity.

One further problem for access deserves mention here.  It sometimes
happens that an ostensibly available resource turns out not to be
available after all.  One may discover the resource because its
creator cited it in a manuscript or an annual research report.
Commonly, a linguist wants to derive recognition for the labor
that went into creating primary language documentation, but
does not want to make the materials available to others until deriving
maximum personal benefit.  Two tactics are to cite
unresolved, non-specific intellectual property rights issues, and to
repeatedly promise but to never finally deliver.
Despite its many guises, this problem has two distinguishing features:
someone draws attention to a resource in order to derive credit for it
-- ``parading their riches'' as Mark Liberman (pers.\ comm.)\ has
aptly described it -- and then applies undocumented or inconsistent
restrictions to prevent access.  The result may be frustration that a
needed resource is withheld, leading to wasted effort or a frozen
project, or to suspicion that the resource is defective and so must
be protected by a smoke screen.

\subsection{Citation}

Research publications are normally required to provide full
bibliographic citation of the materials used in conducting the
research.  Citation standards are high for conventional resources
(such as other publications), but are much lower for language
resources which are usually incorrectly cited, or not cited at all.
This makes it difficult to find out what resource was used in
conducting the research and, in the reverse direction, it is
impossible to use a citation index to discover all the ways in which a
given resource has been applied.

Often a language resource is available on the web, and it is
convenient to have the uniform resource locater (\textsc{url}) since this may
offer the most efficient way to obtain the resource.  However, \textsc{url}s
can fail as a persistent citation in two ways: they may simply break,
or they may cease to reference the same item.  \textsc{url}s break when the
resource is moved or when some piece of the supporting
infrastructure, such as a database server, ceases to work.  Even if a
\textsc{url} does not break, the item it references may be mutable, changing
over time.  Language resources published on the web are usually not
versioned, and a third-party description of some item may cease to be
valid if that item is changed.  Publishing a digital artefact, such as
a \textsc{cd}, with a unique identifier, such as an \textsc{isbn},
avoids this problem.

Citation goes beyond bibliographic citation of a complete item.
We may want to cite some component of a resource, such as a
specific narrative or lexical entry.  However, the format may
not support durable citations to internal components.  For instance, if
a lexical entry is cited by a \textsc{url} which incorporates its lemma, and
if the spelling of the lemma is altered, then the \textsc{url} will not track
the change.
In sum, language documentation and description is not portable if the
incoming and outgoing links to related materials are fragile.

\subsection{Preservation}

The digital technologies used in language documentation and description
greatly enhance our ability to create data
while simultaneously compromising our ability to preserve it.  Relative to paper copy
which can survive for hundreds of years, digitized materials are
evanescent because they use some combination of binary formats with
undocumented character encodings saved on non-archival media and
physically stored with no ongoing administration for backups and
migration to new media.  Presentational markup with \textsc{html} and interactive
content with Javascript and specialized browser plugins require
future browsers to be backwards-compatible.  Furthermore, primary documentation may
be embodied in the interactive behavior of the resource (e.g.\ the gloss
of the text under the mouse may show up in the browser status line,
using the Javascript ``mouseover'' effect).
Consequently, digital resources -- especially dynamic or interactive ones --
often have a short lifespan, and typically become unusable 3-5 years after
they are actively maintained.

\subsection{Rights}

A variety of individuals and institutions may have intellectual
property vested in a language resource,
and there is a complex terrain of legal, ethical and policy issues
\cite{Liberman00}.  In spite of this, most digital language data is
disseminated without identifying the copyright holder and without any
license delimiting the range of acceptable uses of the material.
Often people collect or redistribute materials, or create derived
works without securing the necessary permissions.  While this is
often benign (e.g.\ when the resources are used for research
purposes only), the researcher risks legal action, or having to
restrict publication, or even having to destroy primary materials.  To
avoid any risk one must avoid using materials whose property rights
are in doubt.  In this way, the lack of documented rights restrict the
portability of the language resource.

Sometimes resources are not made available on the web for fear that
they will get into the wrong hands or be misused.  However, this
confuses medium with rights.  The web supports secure data exchange
between authenticated parties (through data encryption) and copyright
statements together with licenses can be used to restrict uses.
More sophisticated models for managing digital rights are
emerging \cite{Iannella01}.  The application of these techniques
to language resources is unexplored, and we are left with an
all-or-nothing situation, in which the existence of any restriction
prevents access across the board.

\subsection{Special challenges for little-studied languages}

Many of the problems reported above also apply to little-studied
languages, though some are greatly exacerbated in this context.  The
small amount of existing work on the language and the concomitant lack
of established documentary practices and conventions may lead to
especially diverse nomenclature.  Inconsistencies within or between
language descriptions may be harder to resolve because of the lack of
significant documentation, the limited access to speakers of the
language, and the limited understanding of dialect variation.  Open
questions in one area of description (e.g.\ the inventory of vowel
phonemes) may multiply the indeterminacies in another (e.g.\ for
transcribed texts).  More fundamentally, existing documentation and
description may be virtually impossible to discover and access, owing
to its fragmentary nature.

The acuteness of these portability problems for little-studied languages
can be highlighted by comparison with well-studied languages.  In English,
published dictionaries and grammars exist to suit all conceivable tastes,
and it therefore matters little (relatively speaking) if none of these
resources is especially portable.  However, when there is only one
dictionary for the language, it must be pressed into a great range of
services, and significant benefits will come from maximizing portability.
\vspace{1ex}

This concludes our discussion of portability problems arising from the way
new tools and technologies are being used in language
documentation and description.  The rest of this paper responds to these
problems, by laying out the core values that
lead to requirements for best practices (\S4) and by providing
best practice recommendations (\S5).
\pagebreak

\section{Value Statements}

Best practice recommendations amount to a decision about which of
several possible options is best.  The notion of best always involves
a value judgment.  Therefore, before making our recommendations,
we articulate the values which motivate our choices.  Our use of
``we'' is meant to include the reader and the wider language resources
community who share these values.

\subsection{Content}

\textsc{Terminology.}
We value the ability of users to identify the substantive similarities and
differences between two resources.
Thus the best practice is one that makes it easy to associate the
comparable parts of unrelated resources.

\textsc{Accountability.}
We value the ability of researchers to verify language descriptions.
Thus the best practice is one that provides the documentation that lies
behind the description.

\textsc{Richness.}
We value the documentation of little-studied languages.
Thus the best practice is one that establishes a record
that is sufficiently broad in scope and rich in detail
that future generations can experience and study the language,
even when no speakers remain.

\subsection{Format}

\textsc{Openness.}
We value the ability of any potential user to make use of a language
resource without needing to obtain unique or proprietary software.
Thus the best practice is one that puts data into a format that is not
proprietary.

\textsc{Documentation.}
We value the ability of potential users of a language resource
to understand its internal structure and organization.
Thus the best practice is one that puts data into a format that
is documented.

\textsc{Machine-Readable.}
We value the ability of users of a language
resource to write programs to process the resource.
Thus the best practice is one that puts the resource into a well-defined format
which can be submitted to automatic validation.

\textsc{Human-Readable.}
We value the ability of users of a language
resource to browse the content of the resource.
Thus the best practice is one that provides a human-digestible
version of a resource.

\subsection{Discovery}

\textsc{Existence.}
We value the ability of any potential user of a language resource to
learn of its existence.
Thus the best practice is one that makes it easy for anyone to discover
that a resource exists.

\textsc{Relevance.}
We value the ability of potential users of a language resource to
judge its relevance without first having to obtain a copy.
Thus the best practice is one that makes it easy for anyone to judge the
relevance of a resource based on its metadescription.

\subsection{Access}

\textsc{Complete.}
We value the ability of any potential user of a language resource to
access the complete resource, not just a limited interface to the resource.
Thus the best practice is one that makes it easy for anyone to obtain
the entire resource.

\textsc{Unimpeded.}
We value the ability of any potential user of a language resource
to follow a well-defined procedure to obtain a copy of the resource.
Thus the best practice is one in which all available resources have
a clearly documented method by which they may be obtained.

\textsc{Universal.}
We value the ability of potential users to access a language
resource from whatever location they are in.
Thus the best practice is one that makes it possible for users to access
some version of the resource regardless of physical location and access
to computational infrastructure.

\subsection{Citation}

\textsc{Credit.}
We value the ability of researchers to be properly credited for the
language resources they create.
Thus the best practice is one that makes it easy for authors to
correctly cite the resources they use.

\textsc{Provenance.}
We value the ability of potential users of a language resource
to know the provenance of the resources it is based on.
Thus the best practice is one that permits resource users to navigate
a path of citations back to the primary linguistic documentation.

\textsc{Persistence.}
We value the ability of language resource creators to endow their work with
a permanent digital identifier which resolves to an instance of the resource.
Thus the best practice is one that associates resources with
persistent digital identifiers.

\textsc{Immutability.}
We value the ability of potential users to cite a language resource without
that resource changing and invalidating the citation.
Thus the best practice is one that makes it easy for authors to
freeze and version their resources.

\textsc{Components.}
We value the ability of potential users to cite the component parts
of a language resource.
Thus the best practice is one that ensures each sub-item of a resource
has a durable identifier.

\subsection{Preservation}

\textsc{Long-Term.}
We value access to language resources over the very long term.
Thus the best practice is one which ensures that language resources
will still be usable many generations into the future.

\textsc{Complete.}
We value the ability of future users of a language resource to
access the complete resource as experienced by contemporary users.
Thus the best practice is one which preserves fragile aspects of
a resource (such as dynamic and interactive content) in a durable
form.

\subsection{Rights}

\textsc{Documentation.}
We value the ability of potential users of a language resource to
know the restrictions on permissible uses of the resource.
Thus the best practice is one that ensures that potential users know
exactly what they are able to do with any available resource.

\textsc{Research.}
We value the ability of potential users of a language resource
to use it in personal scholarship and academic publication.
Thus the best practice is one that ensures that the terms of use on
resources do not hinder individual study and academic research.

\section{Best Practice Recommendations}

This section recommends best practices in support of the values set
out in \S4.  We believe that
the task of identifying and adopting best practices rests with the
community, and we believe that \textsc{olac},
the \textit{Open Language Archives Community},
provides the necessary infrastructure for identifying community-agreed
best practices.  Here, however, we shall attempt to give some broad guidelines
to be fleshed out in more detail later, by ourselves and also, we hope,
by other members of the language resources community.

\subsection{Content}

\textsc{Terminology.}
Map linguistic terminology and descriptive markup
elements to a common ontology of linguistic terms.  This applies to
the obvious candidates such as morphosyntactic abbreviations and
structural markup, but also to less obvious cases such as the
phonological description of the symbols used in transcription.
(NB vocabularies can be versioned and archived in an \textsc{olac} archive;
archived descriptions cite their vocabularies using the \smtt{Relation} element.)

\textsc{Accountability.}
Provide the full documentation on which language descriptions are based.
For example, where a narrative is transcribed, provide the primary recording
(without segmenting it into multiple sound clips).
Create time-aligned transcriptions to facilitate verification.

\textsc{Richness.}
Make rich records of rich interactions, especially in the case of
endangered languages or genres.  Document the ``multimedia linguistic
field methods'' that were used.  Provide theoretically
neutral descriptions of a wide range of linguistic phenomena.

\subsection{Format}

\textsc{Openness.}
Store all language documentation and description in an open format.
Prefer formats supported by multiple third-party software
tools.  NB some proprietary formats are open, e.g.\ Adobe Portable
Document Format (PDF) and MPEG-1 Audio Layer 3 (MP3).

\textsc{Documentation.}
Provide all language documentation and description in a
self-describing format (preferably \textsc{xml}).  Provide detailed documentation of
the structure and organization of the format.  
Encode the characters with Unicode. Try to avoid Private Use Area
characters, but if they are used document them fully.
Document any 8-bit character encodings.
(\textsc{olac} will be providing detailed guidelines for documenting non-standard
character encodings.)

\textsc{Machine-Readable.}
Use open standards such as \textsc{xml} and Unicode,
along with Document Type Definitions (\textsc{dtd}s),
\textsc{xml} Schemas and/or other
definitions of well-formedness which can be verified automatically.
Archive the format definition, giving each version its own unique identifier.
When archiving data in a given format, reference the archived definition
of that format.
Avoid freeform editors for structured information
(e.g.\ prefer Excel or Shoebox over Word for storing lexicons).

\textsc{Human-Readable.}
Provide one or more human readable version of the material,
using presentational markup (e.g.\ \textsc{html}) and/or other convenient
formats.  Proprietary formats are acceptable for delivery
as long as the primary documentation is stored in
a non-proprietary format.

N.B. Format is a critical area for the definition of best practices.
We propose that recommendations in this area be organized by type
(e.g.\ audio, image, text), possibly following the inventory of
types identified in the Dublin Core metadata set.\footnote{%
\url{http://dublincore.org/}}

\subsection{Discovery}

\textsc{Existence.}
List all language resources with an \textsc{olac} data provider.
Any resource presented in \textsc{html} on the web should contain
metadata with keywords and description for use by conventional search
engines.

\textsc{Relevance.}
Follow the \textsc{olac} recommendations on best practice for metadescription,
especially concerning language identification and linguistic data type.
This will ensure the highest possibility of discovery by interested users
in the \textsc{olac} union catalog hosted by Linguist.\footnote{%
\url{http://www.linguistlist.org/}}

\subsection{Access}

\textsc{Complete.}
Publish complete primary documentation.
Publish the documentation itself, and not just an interface to
it, such as a web search form.

\textsc{Unimpeded.}
Document all access methods and restrictions along with other metadescription.
Document charges and expected delivery time.

\textsc{Universal.}
Make all resources accessible by any interested user.
Publish digital resources using appropriate delivery media,
e.g.\ web for small resources, and \textsc{cd/dvd} for large resources.
Where appropriate, publish corresponding print versions,
e.g.\ for the dictionary of a little-studied language.

\subsection{Citation}

\textsc{Credit, Provenance.}
Furnish complete bibliographic data for all language resources created.
Provide complete citations for all language resources used.
Document the relationship between resources in the
metadescription (NB in the \textsc{olac} context, use the \smtt{Relation} element).

\textsc{Persistence.}
Ensure that resources have a persistent identifier, such as an \textsc{isbn}
or a persistent \textsc{url} (e.g.\ a Digital Object Identifier\footnote{%
\url{http://www.doi.org/}}).
Ensure that at least one persistent identifier resolves to an instance of
the resource or to detailed information about how to obtain the resource.

\textsc{Immutability.}
Provide fixed versions of a resource, either by publishing it on a
read-only medium, and/or submitting it to an archive which ensures
immutability.  Distinguish multiple versions with a version number or
date, and assign a distinct identifier to each version.

\textsc{Components.}
Provide a formal means by which the components of a resource may be
uniquely identified.  Take special care to avoid the possibility of
ambiguity, such as arises when lemmas are used to identify lexical
entries, and where multiple entries can have the same lemma.

\subsection{Preservation}

\textsc{Long-Term.}
Commit all documentation and description to a digital
archive which can credibly promise long-term preservation and access.
Ensure that the archive satisfies the key requirements of a well-founded
digital archive (e.g.\ implements
digital archiving standards, provides offsite backup, migrates materials
to new formats and media/devices over time, is committed to supporting new
access modes and delivery formats, has long-term institutional support,
and has an agreement with a national archive to take materials
if the archive folds).
Archive physical versions of the language documentation and description
(e.g.\ printed versions of documents; any tapes from which
online materials were created).
Archive electronic documents using type 1 (scalable) fonts in preference
to bitmap fonts.

\textsc{Complete.}
Ensure that all aspects of language documentation and description
accessible today are accessible in future.  Ensure that any
documentary information conveyed via dynamic or interactive behaviors
is preserved in a purely declarative form.

\subsection{Rights}

\textsc{Documentation.}
Ensure that the intellectual property rights relating to the resource are
fully documented.

\textsc{Research.}
Ensure that the resource may be used for research purposes.

\section{Conclusion}

Today, the community of scholars engaged in language documentation and
description exists in a cross-over period between the paper-based era
and the digital era.  We are still working out how to preserve knowledge
that is stored in digital form.  During this transition period, we
observe unparalleled confusion in the management of digital language
documentation and description.  A substantial fraction
of the resources being created can only be re-used on the same software/hardware
platform, within the same scholarly community, for the same purpose, and then only
for a period of a few years.  However, by adopting a range of best practices, this
specter of chaos can be replaced with the promise of easy access to
highly portable resources.

Using tools as our starting point, we described a diverse range of practices
and discussed their negative implications for data portability along seven dimensions,
leading to a collection of advice for how to create portable resources.  These
three categories, tools, data, and advice, are three pillars of the infrastructure
provided by \textsc{olac}, the Open Language Archives Community
\cite{BirdSimons01}.  Our best practice recommendations are preliminary, and
we hope they will be fleshed out by the community using the \textsc{olac}
Process.\footnote{%
\url{http://www.language-archives.org/OLAC/process.html}}

We leave off where we began, namely with tools.  It is our use of the new
tools which have led to data portability problems.  And it is only
with new tools, supporting the kinds of best practices we recommend,
which will address these problems.  An archival format is useless
unless there are tools for creating, managing and browsing the content
stored in that format.  Needless to say, no single organization has
the resources to create the necessary tools, and no third party
developing general-purpose office software will address the unique
needs of the language documentation and description community.
We need nothing short of an open source revolution, leading to
new specialized tools based on shared data models for all of the
basic linguistic types, and connected to portable data formats.

\section*{Acknowledgements}

This research is supported by NSF
Grant No.\ 9983258 ``Linguistic Exploration'' and
Grant No.\ 9910603 ``International Standards in Language Engineering.''

\renewcommand{\enotesize}{\scriptsize}
\theendnotes

\scriptsize\raggedright
\bibliographystyle{lrec2000}

\end{document}